\pdfoutput=1

\documentclass[11pt]{article}

\usepackage[preprint]{acl}

\usepackage{times}
\usepackage{latexsym}
\usepackage{amsmath}
\usepackage{amssymb}
\usepackage{txfonts}
\usepackage{graphicx}
\usepackage{booktabs}
\usepackage{subcaption}
\usepackage{enumitem}
\usepackage{makecell}

\usepackage[T1]{fontenc}

\usepackage[utf8]{inputenc}

\usepackage{microtype}

\usepackage{inconsolata}

\usepackage{graphicx}

\newcommand{\condprob}[2]{\mathsf{P}( #1 \mid #2 )}

\newcommand{\magenta}[1] {\textcolor{magenta}{#1}}

\title{The Impact of Token Granularity on the Predictive Power of \\ Language Model Surprisal}

\author{Byung-Doh Oh \\
  Center for Data Science \\
  New York University \\
  \texttt{oh.b@nyu.edu} \\\And
  William Schuler \\
  Department of Linguistics \\
  The Ohio State University \\
  \texttt{schuler.77@osu.edu}}

\begin{document}
\maketitle
\begin{abstract}
Word-by-word language model surprisal is often used to model the incremental processing of human readers, which raises questions about how various choices in language modeling influence its predictive power.
One factor that has been overlooked in cognitive modeling is the granularity of subword tokens, which explicitly encodes information about word length and frequency, and ultimately influences the quality of vector representations that are learned.
This paper presents experiments that manipulate the token granularity and evaluate its impact on the ability of surprisal to account for processing difficulty of naturalistic text and garden-path constructions.
Experiments with naturalistic reading times reveal a substantial influence of token granularity on surprisal, with tokens defined by a vocabulary size of 8,000 resulting in surprisal that is most predictive.
In contrast, on garden-path constructions, language models trained on coarser-grained tokens generally assigned higher surprisal to critical regions, suggesting a greater sensitivity to garden-path effects than previously reported.
Taken together, these results suggest a large role of token granularity on the quality of language model surprisal for cognitive modeling.
\end{abstract}

\section{Introduction}
In cognitive modeling, word-by-word surprisal is often used as a predictor of processing difficulty, under a theoretical framework that emphasizes the predictive aspect of real-time language processing \citep{hale01, levy08}.
In recent years, neural network-based language models (LMs) have been used to calculate and evaluate surprisal against human reading times \citep{wilcoxetal20, merkxfrank21}, which has opened possibilities for refining them as computational models of language processing and using them to study how predictive processing interacts with other cognitive processes.
Therefore, a core question in this area is how various aspects of language modeling such as the LMs' architecture or training data influence the learned probabilities and their alignment to human-like processing difficulty.

\begin{figure}[t!]
    \centering
    \footnotesize
    `If you were to journey'

    \vspace{1em}
    \textbf{Finer granularity, more character-like} ($|V|=256$)
    
    \setlength\fboxsep{1pt}
    \vspace{0.25em}
    \texttt{\colorbox{lightgray!75}{␣} \colorbox{lightgray!75}{I} \colorbox{lightgray!75}{f} \colorbox{lightgray!75}{␣} \colorbox{lightgray!75}{y} \colorbox{lightgray!75}{o} \colorbox{lightgray!75}{u} \colorbox{lightgray!75}{␣w} \colorbox{lightgray!75}{er} \colorbox{lightgray!75}{e} \colorbox{lightgray!75}{␣to} \colorbox{lightgray!75}{␣} \colorbox{lightgray!75}{j} \colorbox{lightgray!75}{o} \colorbox{lightgray!75}{ur} \colorbox{lightgray!75}{n} \colorbox{lightgray!75}{e} \colorbox{lightgray!75}{y}}

    \vspace{1em}
    \textbf{Coarser granularity, more word-like} ($|V|=128000$)

    \vspace{0.25em}
    \texttt{\colorbox{lightgray!75}{␣If} \colorbox{lightgray!75}{␣you} \colorbox{lightgray!75}{␣were} \colorbox{lightgray!75}{␣to} \colorbox{lightgray!75}{␣journey}}

    \caption{Smaller subword vocabulary sizes result in longer sequences of finer-granularity \texttt{\colorbox{lightgray!75}{tokens}} that are more character-like (top), and larger vocabulary sizes result in shorter sequences of coarser-granularity \texttt{\colorbox{lightgray!75}{tokens}} that are more word-like (bottom).}
    \label{fig:overview}
\end{figure} 

One such variable that has been overlooked in cognitive modeling is the granularity of the tokens over which LMs are trained to define probability distributions.
To allow LMs to flexibly handle unseen word forms and keep the vocabulary size manageable, it has become a standard practice in language modeling to use `subword' tokenizers \citep[e.g.][]{sennrichetal15, kudo18}.
Such tokenizers are often trained on corpus statistics such that their vocabularies contain a fixed number of frequent sequences (which may or may not correspond to words) as independent tokens.
As a consequence, less frequent word forms are split into multiple subword tokens during both training and inference.

This suggests that there are at least two ways in which token granularity will influence the quality of LM surprisal in the context of cognitive modeling.
The first is through the different initial biases about word probabilities that the different levels of token granularity embody.
For example, a tokenizer with a very fine token granularity (Figure \ref{fig:overview}, top) tokenizes the word \textit{journey} into seven tokens, and therefore a uniform initial distribution over tokens predicts its probability to be six magnitudes lower than that of \textit{to}.
In contrast, a tokenizer with coarser granularity (Figure \ref{fig:overview}, bottom) keeps both \textit{journey} and \textit{to} intact, and a uniform initial distribution over these tokens predicts their probabilities to be equal.
As such subword tokenization is informed by word length and frequency, which are variables that are well known to influence real-time processing \citep{bartonetal14, justcarpenter80}, some tokenization schemes are more likely to lead to word-level surprisal that is more predictive of processing difficulty.

The second way in which token granularity will impact LM surprisal is through the quality of the token representations that are learned.
That is, LMs trained on coarse-grained tokens that are more word-like will learn token representations that align closer to lexical co-occurrence statistics like those from earlier neural network LMs \citep{mikolovetal13nips}.
In contrast, fine-grained tokens will cause words to be split across multiple vector representations, which may make learning word-to-word associations more challenging.
For instance, the LM would need to attend to seven separate vector representations for \textit{journey} to predict \textit{travel} later in the sequence.

Motivated by these observations, this work presents experiments that manipulate the token granularity of LMs and evaluate its influence on the ability of surprisal to account for processing difficulty of both naturalistic text and garden-path constructions.
First, regression experiments across multiple reading time corpora reveal a strong influence of token granularity on the predictive power of surprisal prior to any LM training, with tokens defined by a vocabulary size of 4,000 resulting in surprisal that is most predictive.
This initial bias appears to persist throughout training in smaller LMs, and eventually yield surprisal that is more predictive than that of larger LMs.
In contrast, LMs trained with tokens of coarser granularity generally assigned higher surprisal to critical regions of garden-path constructions, suggesting a greater sensitivity to garden-path effects.
These results suggest a large role of token granularity on the quality of LM surprisal for cognitive modeling, with different levels of granularity being more appropriate for modeling broad-coverage comprehension and garden-path effects.

\section{Related Work}
The choice of subword tokenizers has mostly been studied in terms of performance on natural language processing tasks \citep[e.g.][]{bostromdurrett20}, with more recent work aiming to provide explanations for their downstream impact \citep{zouharetal23, schmidtetal24} or the discrepancy in performance across languages \citep{arnettbergen25}.
Other work proposes methods for mitigating `quirks' introduced by subword tokenization, such as LMs' sensitivity to misspelling \citep{vieiraetal24} or `under-trained' tokens with poor quality representations \citep{landbartolo24}. 

In psycholinguistic modeling, \citet{ohetal21acl} incorporate a character model to an incremental left-corner parser and show improvements in surprisal's fit to naturalistic reading times.
\citet{nairresnik23} compare a subword tokenization scheme informed by concatenative morphology against a word-level and a byte-pair encoding scheme and report its impact on LM surprisal's predictions of reading times.
\citet{giulianellietal24} apply \citeauthor{vieiraetal24}'s \citeyearpar{vieiraetal24} algorithm to derive character-level probabilities from GPT-2's \citep{radfordetal19} token-level probabilities and demonstrate the potential of using them to model the processing difficulty of specific focal areas (e.g.~the first three characters of a word) more flexibly.

\section{Experiment 1: Impact on Fit to Naturalistic Reading Times} \label{sec:exp1}
The first experiment evaluates the effect of token granularity on the fit of LM surprisal to naturalistic reading times.
LM surprisal is first evaluated prior to any training to examine the extent to which word-level surprisal determined purely by token granularity is predictive of reading times.
Subsequently, surprisal from LMs of different sizes is evaluated after training to examine how model size and training data interact with these initial biases.

\subsection{Vocabulary Construction} \label{sec:vocab}
We manipulate the granularity of the LM tokens by training subword tokenizers with vocabularies of different sizes, upon which the LM training and surprisal calculation is based.

\paragraph{Subword Tokenizer.} We used the unigram language model tokenizer \citep[ULM;][]{kudo18} to construct subword vocabularies of different sizes.
The ULM tokenizer aims to find a vocabulary $V$ of a desired size that maximizes the joint probability of the subword sequence.
This is achieved by starting with a large vocabulary and then iteratively pruning entries that result in the smallest drop in the marginal likelihood over possible tokenizations.
The ULM tokenizer is similar to the byte-pair encoding tokenizer \citep[BPE;][]{sennrichetal15} in that it `compresses' frequent subword sequences into tokens, but different in that it treats the character as the basic subword unit (cf.~bytes as basic units in BPE\footnote{As bytes are less interpretable than characters, the ULM tokenizer was opted for in this work.}) and defines a probability distribution over possible tokenizations given a string (cf.~BPE is deterministic).

\paragraph{Training Data.} The ULM tokenizer requires a training corpus for estimating the probability of each subword token and ultimately constructing the vocabulary.
To this end, we used a subset of the English training section of the Wiki-40B dataset \citep{guoetal20}, which contains Wikipedia articles.
The Wiki-40B dataset was chosen as it closely approximates the domain represented by the reading time corpora that are of interest, in contrast to other pre-training corpora that also include e.g.~programming code.
After removing the metadata tags and filtering articles that are not in English, 1,000,000 articles were sampled to train the ULM tokenizer.

\paragraph{Training Setup.}
Subsequently, ULM tokenizers were trained to vocabulary sizes of \{256, 512, 1000, 2000, 4000, 8000, 16000, 32000, 48000, 64000, 128000\} to cover a wide range of token granularity.\footnote{LMs widely used in cognitive modeling like GPT-2 \citep{radfordetal19} or Pythia \citep{bidermanetal23} have a vocabulary size of around 50,000.}
The character-level coverage of the training corpus was set to 99.95\%, which resulted in a set of 158 characters as the basic subword units.
The final vocabulary of each condition consists of these basic subword units, frequent subword sequences determined by the training process, an \texttt{<unk>} token for characters that are not in the set of basic subword units, and an \texttt{<s>} token for denoting the start of sequence.

\subsection{Language Modeling} \label{sec:lm}
Based on the tokenizers trained following the procedures in Section \ref{sec:vocab}, autoregressive LMs with different vocabulary sizes were subsequently trained.

\paragraph{Model Architecture.}
As can be seen in Figure~\ref{fig:overview}, different token granularities result in token sequences of varying lengths over the same text string.
This poses a stark challenge for training Transformers \citep{vaswanietal17transformer} with different vocabulary sizes, whose self-attention mechanism has unfavorable space complexity for long input sequences.\footnote{The conventional solution to this issue is to set a maximum input sequence length, which would nonetheless result in LMs that condition on different amounts of text depending on token granularity.}
To overcome this issue, we train LMs based on the Mamba-2 architecture \citep{daogu24}, which belongs to the class of state space models (SSMs).
Each Mamba-2 block aggregates representations from previous timesteps through recurrence-like operations defined by timestep-specific parameters that are determined by the input.
Similarly to Transformer's attention, Mamba-2's operations can be distributed across multiple `heads' that have different parameters.
For the experiments described in this paper, we use the multi-input setup of Mamba-2, which was shown to be the most effective for language modeling \citep{daogu24}.\footnote{The multi-input setup is analogous to multi-value attention, where different attention heads share the query and key matrices but not the value matrices.}

To examine the potential impact of the LM's size on the fit of surprisal to reading times \citep{ohschuler23tacl, shainetal24}, we trained models of three different sizes for each tokenizer condition.
Following conventional practice, we varied the number of layers,\footnote{Because a Mamba-2 layer does not contain a multi-layer perceptron like a Transformer layer, \citet{daogu24} argue that a Mamba-2 model needs twice as many layers in order to be comparable with a Transformer model.} the number of `SSM heads' on each layer, and the model embedding size.
Additionally, following \citet{daogu24}, the state size was set to twice the model embedding size, and the size of each SSM head was fixed to 64.
Finally, the token embeddings were shared across the initial input and the final projection layers.
The resulting hyperparameters and the total number of non-embedding parameters are summarized in Table \ref{tab:params}.

\begin{table}[t!]
    \centering
    \begin{tabular}{lrrrr} \toprule
    Model & \#L & \#H & $d_{\text{model}}$ & \#Parameters \\ \midrule
    \textit{Small} & 6 & 8 & 256 & 2,592,400 \\
    \textit{Medium} & 12 & 16 & 512 & 19,847,744 \\
    \textit{Large} & 24 & 24 & 768 & 87,993,792 \\ \bottomrule
    \end{tabular}
    \caption{Hyperparameters of LMs that were trained in this work. \#L, \#H, and $d_{\text{model}}$ refer to number of Mamba-2 layers, number of SSM heads per layer, and model embedding size, respectively. \#Parameters exclude parameters from the token embeddings, the number of which are different across each tokenizer.}
    \label{tab:params}
\end{table}

\paragraph{Training Data.}
We used the entire English training section of the Wiki-40B dataset \citep{guoetal20} to train the LMs.
Following similar procedures as the tokenizer training, metadata tags and articles that are not in English were removed.
Moreover, one article that overlapped substantially in content with the reading time corpora was also filtered out.
Each remaining article was then treated as a single training example for the LMs.
The ULM tokenizers trained in Section \ref{sec:vocab} were used to tokenize the articles by returning the most likely tokenization of the string.

While the Mamba-2 architecture is more favorable toward longer sequences, some articles were excessively long when tokenized into finer-grained tokens.
Therefore, a small subset of long articles was further split into `Wikipedia sections' to alleviate this issue.
This resulted in a total of 5,152,219 training examples.

\paragraph{Training Setup.}
One iteration of the training data was provided in the same order to each LM in 10,063 training batches of 512 examples.
The AdamW optimizer \citep{loshchilovhutter19} with a maximum learning rate of $1 \times 10^{-3}$ was used to train the model parameters.
This maximum learning rate was linearly warmed up over the first $\sim$5\% of training steps (i.e.~503 steps) and was subsequently annealed to a minimum of $1 \times 10^{-5}$ following a cosine schedule over the remaining training steps.
Gradients were clipped to a maximum norm of 1 to further stabilize training.
All LM training took place in half-precision on a 48GB Nvidia RTX 8000 GPU.\footnote{The trained ULM tokenizers and Mamba-2 models are publicly available at \url{https://github.com/byungdoh/ssm-surprisal}.}

\subsection{Reading Time Modeling} \label{sec:rt}
Surprisal from LMs trained following the procedures in Section \ref{sec:lm} was subsequently evaluated on its ability to predict naturalistic reading times.
Following recent work \citep{shain24, shainetal24}, this experiment aimed to identify trends using data across multiple reading time corpora.

\paragraph{Reading Time Corpora.}
The reading time data analyzed in this experiment consist of 10 measures from five self-paced reading (SPR) and eye-tracking (ET) corpora:
\begin{enumerate}[leftmargin=*, itemsep=0em]
    \item Natural Stories \citep{futrelletal21}: SPR times from 181 subjects that read 10 naturalistic English passages of narrative and expository text (10,256 words).
    \item Brown \citep{smithlevy13}: SPR times from 35 subjects that read 13 English passages (7,180 words) from the Brown Corpus \citep{kucerafrancis67}.
    \item GECO \citep{copetal17}: Fixation durations from 14 monolingual subjects that read the English version of novel \textit{The Mysterious Affair at Styles} (\citealp{christie20}; 13 chapters, 56,411 words).
    \item Dundee \citep{kennedyetal03}: Fixation durations from 10 subjects that read 67 English newspaper editorials (51,501 words).
    \item Provo \citep{lukechristianson18}: Fixation durations from 84 subjects that read 55 short English passages (2,746 words) ranging between news articles, science magazines, and fictional work.
\end{enumerate}

\paragraph{Data Preprocessing and Partitioning.}
For the SPR datasets, by-word reading times were filtered to exclude those of words at sentence boundaries (i.e.~sentence-initial and -final) and those shorter than 100 ms or longer than 3,000 ms.
For the Natural Stories Corpus, data from subjects who answered fewer than five comprehension questions correctly were also removed.

\newlength{\oldtabcolsep}
\setlength{\oldtabcolsep}{\tabcolsep}
\setlength{\tabcolsep}{3pt}
\begin{table}[t!]
    \centering
    \footnotesize
    \begin{tabular}{lrrr} \toprule
    Corpus/Measure & Fit & Exploratory & Held-out \\ \midrule
    Natural Stories SPR & 384,984 & 192,826 & 192,449 \\
    Brown SPR & 59,292 & 29,671 & 30,157 \\ \midrule    
    GECO FP & 144,850 & 72,468 & 72,574 \\
    GECO GP & 144,850 & 72,468 & 72,574 \\
    Dundee SP & 155,483 & 77,809 & 77,101 \\
    Dundee FP & 98,115 & 48,598 & 48,794 \\
    Dundee GP & 98,115 & 48,598 & 48,794 \\
    Provo SP & 91,032 & 45,654 & 45,404 \\
    Provo FP & 52,959 & 26,539 & 26,640 \\
    Provo GP & 52,960 & 26,539 & 26,640 \\ \bottomrule
    \end{tabular}
    \caption{Number of data points in each partition of each reading time dataset.}
    \label{tab:observations}
\end{table}
For the ET data that contains non-linear eye movements, the scan path (SP), first-pass (FP), and go-past (GP) durations were calculated and analyzed for each word region.\footnote{The SP duration could not be calculated for the GECO dataset that does not provide raw fixation data.}
These datasets were filtered to remove data for unfixated words, words following saccades longer than four words, and words at sentence and document boundaries.
Data points corresponding to words at line and screen boundaries were also excluded for the Dundee Corpus that provides relevant annotations.

After data preprocessing, each dataset was partitioned into fit, exploratory, and held-out partitions that comprise roughly 50\%, 25\%, and 25\% of data points respectively.
This partitioning was conducted based on the sum of the subject ID and the sentence ID,\footnote{To this end, the sum of the subject ID and sentence ID modulo four was calculated (zero or one: fit partition, two: exploratory partition, three: held-out partition).} which keeps all data from a particular subject reading a particular sentence in one partition.
The fit partition was used to estimate the regression parameters, and all results are reported on the exploratory partition.
The held-out partition is reserved for any statistical significance testing, and its use is kept minimal.
The number of data points after preprocessing and partitioning is outlined in Table \ref{tab:observations}.

\begin{table*}[ht!]
    \centering
    \begin{tabular}{lrrrrrrrrrrr} \toprule
    Model \textbackslash \ $|V|$ & 256 & 512 & 1000 & 2000 & 4000 & 8000 & 16000 & 32000 & 48000 & 64000 & 128000 \\ \midrule
    \textit{Small} & 2205.8 & 2292.7 & 2338.2 & 2386.2 & \textbf{2560.2} & 2531.5 & 2552.0 & 2360.3 & 2220.8 & 2088.6 & 1901.9 \\
    \textit{Medium} & 2237.5 & 2312.7 & 2344.8 & 2393.5 & \textbf{2555.8} & 2541.0 & 2533.7 & 2360.5 & 2210.9 & 2101.5 & 1902.9 \\
    \textit{Large} & 2212.1 & 2274.4 & 2321.6 & 2395.2 & \textbf{2542.4} & 2519.0 & 2536.2 & 2362.8 & 2204.1 & 2049.3 & 1890.9 \\ \midrule
    Average & 2218.5 & 2293.3 & 2334.9 & 2391.6 & \textbf{2552.8} & 2530.5 & 2540.6 & 2361.2 & 2211.9 & 2079.8 & 1898.6 \\ \bottomrule
    \end{tabular}
    \caption{$\Delta$LogLik from LM surprisal prior to any LM training, aggregated over the exploratory partitions of all reading time datasets.}
    \label{tab:exp1_init}
\end{table*}

\paragraph{Surprisal Calculation.}
Each passage of the five reading time corpora was tokenized using each LM's respective ULM tokenizer and provided as input to calculate token probabilities that were converted to word probabilities.
Preliminary analyses showed that the trained ULM tokenizers prepend the whitespace character to tokens, such that they have leading whitespaces.
However, if word probabilities are naively calculated with leading whitespaces (e.g.~$\condprob{\textit{were}}{\textit{If you}}$ calculated as $\condprob{\texttt{\textvisiblespace were}}{\texttt{\textvisiblespace If \textvisiblespace you}}$ in Figure \ref{fig:overview}), the sum over all word probabilities can exceed one, as the tokens do not explicitly mark the end of the word. 
Recent work \citep{ohschuler24, pimentelmeister24} provides a correction for this issue that factors the probability of each whitespace and re-allocates it to its preceding token (i.e.~$\condprob{\textit{were}}{\textit{If you}}$ calculated as $\condprob{\texttt{were \textvisiblespace}}{\texttt{\textvisiblespace If \textvisiblespace you \textvisiblespace}}$), which was applied in this work to calculate word probabilities.

\paragraph{Regression Modeling.} \label{sec:lmer}
We fit linear mixed-effects \citep[LME;][]{batesetal15} regression models using each fit partition of the reading time datasets to evaluate the impact of token granularity on LM surprisal's fit to reading times.
The goodness-of-fit was evaluated by calculating the increase in regression model log-likelihood ($\Delta$LogLik) on each exploratory partition due to including each LM surprisal predictor on top of the baseline regression model.

The baseline regression models contain as predictors word length in characters, index of word position within the sentence, unigram surprisal (all datasets), and whether the previous word was fixated (ET datasets only).
Unigram surprisal was calculated using the KenLM toolkit \citep{heafieldetal13} with parameters estimated on the OpenWebText Corpus \citep{gokaslancohen19}.
On top of these baseline regression models, surprisal of the current word and the preceding word was included to capture spillover effects \citep{rayneretal83}.

The raw reading times were not transformed prior to regression modeling, assuming a linear relationship between surprisal and reading times \citep{wilcoxetal23, xuetal23, shainetal24}.
The LME models were fit with maximal random effects that were supported by the data \citep{barretal13} by removing the least predictive random effect until all LME models converged.
The models fit to SPR data included by-subject random slopes for word position, word length, and surprisal of current and previous word.
The models fit to ET data included by-subject random slopes for word position and surprisal of current word.
All LME models also include a by-subject random intercept.
These regression modeling procedures were repeated for all LMs prior to any LM training (i.e.~at initialization) and at the end of LM training. 
The corpus-level perplexity is also reported at the end of LM training, based on prior results showing a systematic relationship between perplexity and surprisal's fit to reading times \citep{wilcoxetal20, wilcoxetal23qp, ohschuler23emnlp}.

\subsection{Results}
The results from LMs prior to training in Table \ref{tab:exp1_init} reveal a strong influence of token granularity on the predictive power of word-level surprisal, with surprisal from tokenizers with vocabulary sizes of 4,000, 8,000, and 16,000 showing the strongest fits.
As the vocabulary size increases and subword tokens become more word-like, the fit of surprisal to reading times declines as more and more words are similarly assigned `uniform surprisal.'
On the other hand, as the vocabulary size decreases and subword tokens become more character-like, the fit to reading times also declines because surprisal becomes more strongly correlated with word length that is included as a baseline predictor.
The intermediate vocabulary sizes appear to represent an optimum along this continuum.

\begin{figure}[t!]
    \centering
    \includegraphics[width=\linewidth]{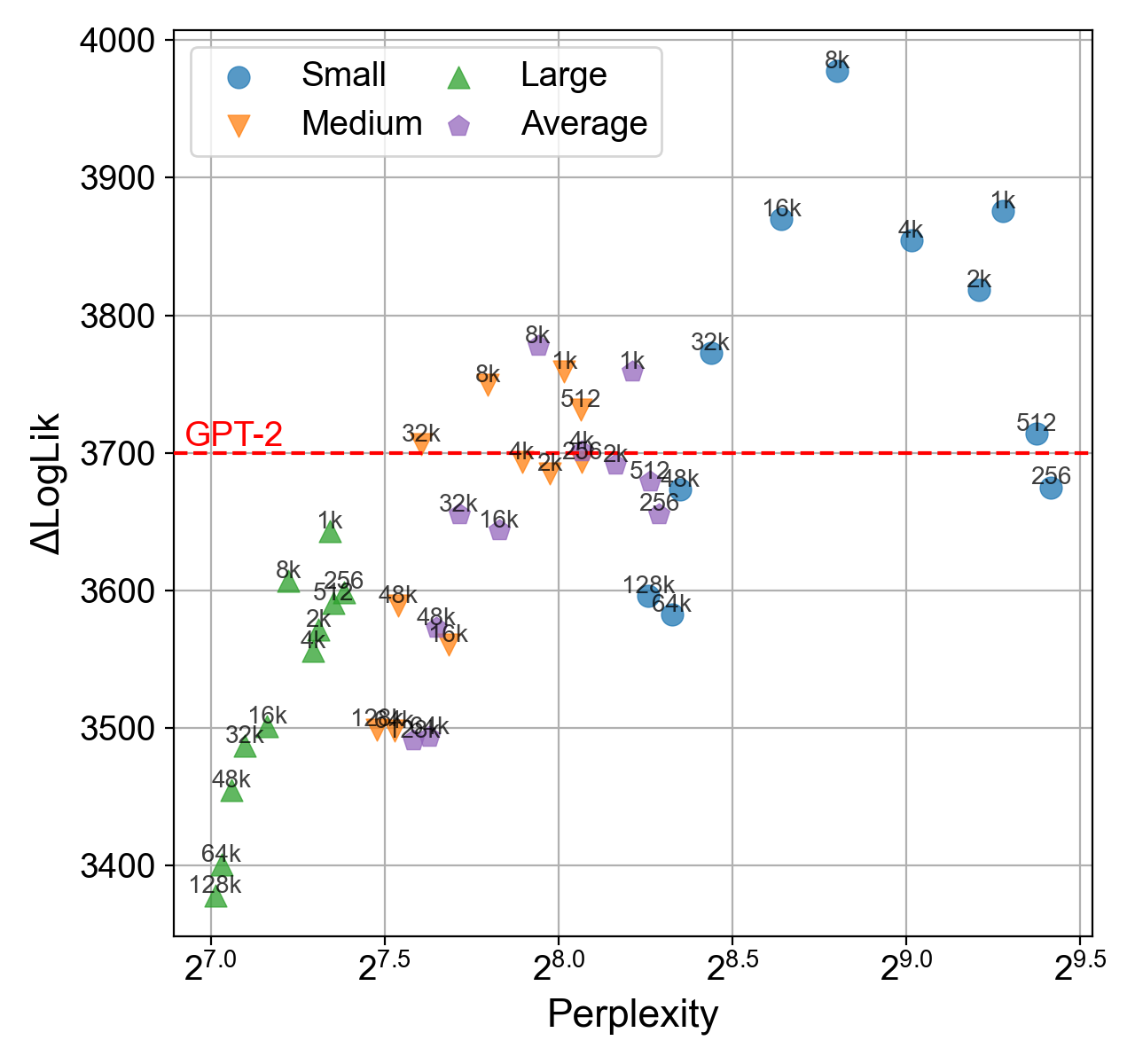}
    \caption{$\Delta$LogLik from LM surprisal on the exploratory partitions and corpus-level perplexity after LM training, both aggregated over all reading time corpora. The `Average' represents the arithmetic mean of $\Delta$LogLik and the geometric mean of perplexity over the three model sizes. The red line denotes aggregate $\Delta$LogLik from GPT-2 Small surprisal \citep{radfordetal19} for reference. See Appendix~\ref{sec:per_corpus} for the results on each individual dataset.}
    \label{fig:exp1_final}
\end{figure}

At the end of LM training, a strong interaction is observed between LM size and token granularity (Figure \ref{fig:exp1_final}).
While surprisal from the \textit{Small} LMs generally replicate the peak observed prior to LM training, this peak becomes less pronounced with the \textit{Medium} and \textit{Large} LMs. 
In particular, the \textit{Large} LMs show much smaller differences in both perplexity and $\Delta$LogLik across different vocabulary sizes.
This suggests that increased model sizes allow LMs to learn qualitatively similar predictions that overcome the initial biases imposed by token granularity.
However, considering the average over model sizes (i.e.~the purple points in Figure \ref{fig:exp1_final}), we conclude that the token granularity represented by a vocabulary size of around 8,000 results in surprisal estimates that are the strongest predictors of naturalistic reading times, even over the widely-used GPT-2 Small \citep{radfordetal19}.

\section{Experiment 2: Impact on Magnitude of Surprisal-Based Garden-Path Effects} \label{sec:exp2}
The second experiment evaluates the effect of token granularity on the magnitude of surprisal-based estimates of garden-path effects (GPE).
The aim of this experiment is to evaluate how token granularity influences the LMs' sensitivity to syntax by evaluating their surprisal on more targeted syntactic constructions.

\setlength{\tabcolsep}{\oldtabcolsep}
\begin{table*}[ht!]
    \centering
    \footnotesize
    \begin{tabular}{ll} \toprule
    Construction/Condition & Example \\ \midrule
    \makecell[l]{MV/RR Ambiguous \\ MV/RR Unambiguous} & \makecell[l]{The suspect sent the file \textit{\magenta{deserved} further investigation} given the new evidence. \\ The suspect who was sent the file \textit{\magenta{deserved} further investigation} given the new evidence.} \\ \midrule
    \makecell[l]{NP/S Ambiguous \\ NP/S Unambiguous} & \makecell[l]{The suspect showed the file \textit{\magenta{deserved} further investigation} during the murder trial. \\ The suspect showed that the file \textit{\magenta{deserved} further investigation} during the murder trial.} \\ \midrule
    \makecell[l]{NP/Z Ambiguous \\ NP/Z Unambiguous} & \makecell[l]{Because the suspect changed the file \textit{\magenta{deserved} further investigation} during the jury discussions. \\ Because the suspect changed, the file \textit{\magenta{deserved} further investigation} during the jury discussions.} \\ \bottomrule
    \end{tabular}
    \caption{Examples of garden-path constructions studied in \citet{huangetal24}. In each sentence pair, the critical word is highlighted in magenta, and its two spillover words are italicized. In the ambiguous conditions, the critical word disambiguates the syntactic structure of the sentence and incurs processing difficulty.}
    \label{tab:gp_constructions}
\end{table*}

\begin{figure*}[ht!]
    \centering
    \includegraphics[width=\textwidth]{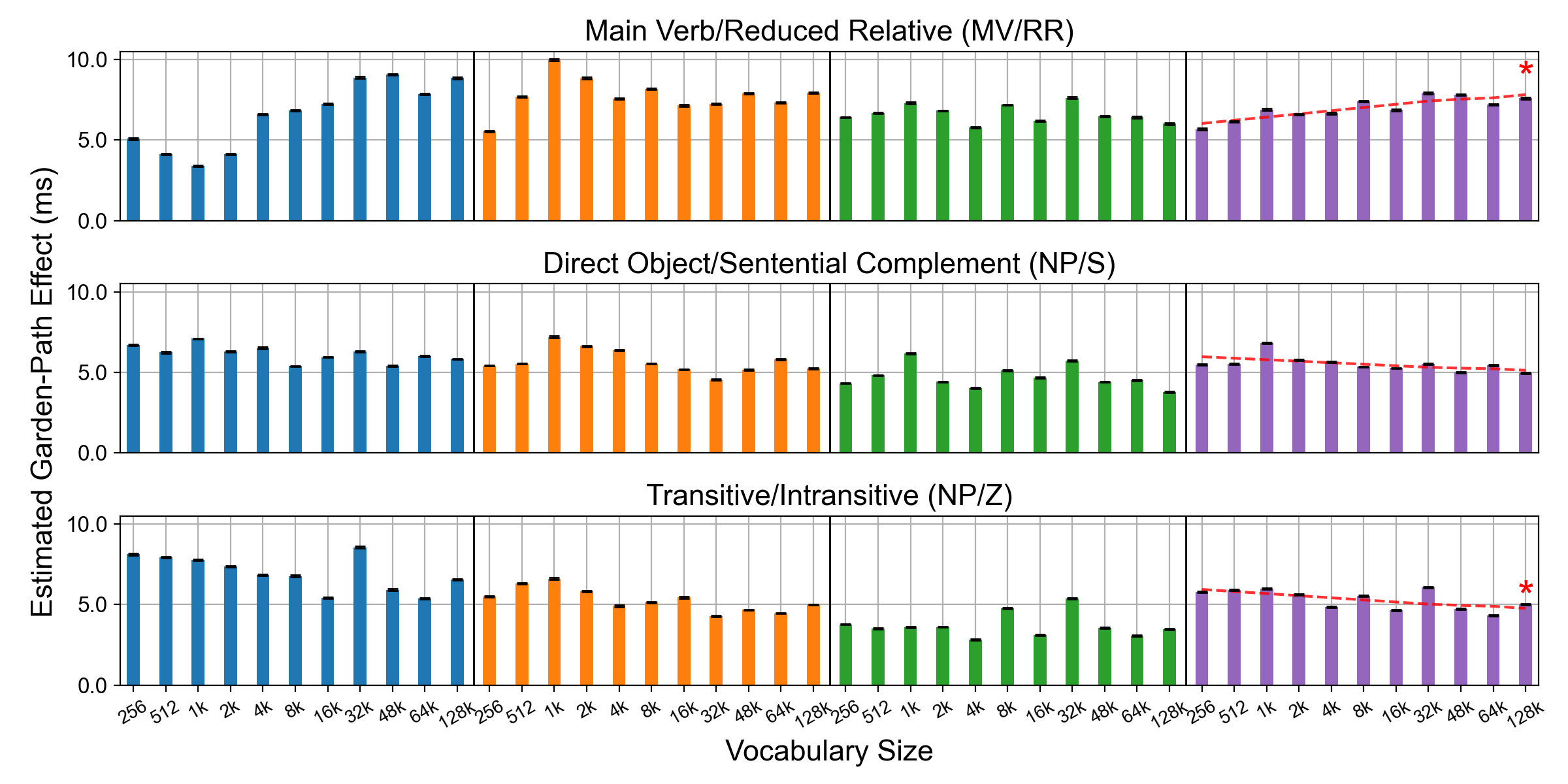}
    \includegraphics[width=0.67\textwidth]{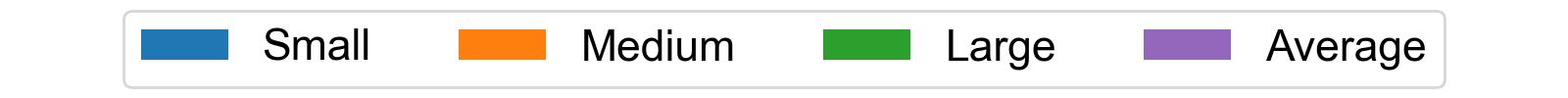}
    \caption{Estimated garden-path effects at the first spillover word (\textit{further} in Table \ref{tab:gp_constructions}) for the three garden-path constructions using LM surprisal. Error bars denote 95\% confidence intervals. The red curve shows predictions from the line of best fit with $\log_2(|V|)$ as the independent variable. $*$: $p < 0.05$.}
\label{fig:gpe_p1}
\end{figure*}

\subsection{Procedures}
We estimated surprisal-based GPE from the LMs trained in Section \ref{sec:lm}, using the data and following the modeling procedures of \citet{huangetal24}.

\paragraph{Surprisal-RT Linking Function.}
First, to estimate a linking function between LM surprisal and human reading times, LME models were fit to raw SPR times ($n{=}995,814$) of filler items (i.e.~`ordinary' sentences that do not incur processing difficulty due to syntactic disambiguation) drawn from the Provo Corpus \citep{lukechristianson18} that are provided by \citet{huangetal24}.
These filler LME models include LM surprisal and log frequency of the current word and two previous words, word length in characters, and index of word position within the sentence as main effects, as well as by-subject and by-item random intercepts.
These modeling choices make similar assumptions about the functional form between surprisal and reading times and the lingering influence of previous words as Experiment 1.

\paragraph{Garden-Path Stimuli and Reading Time Data.}
The stimuli used in \citet{huangetal24} consist of 24 items of the Main Verb/Reduced Relative (MV/RR), Direct Object/Sentential Complement (NP/S), and Transitive/Intransitive (NP/Z) garden-path constructions.
Each item consists of a sentence in the ambiguous condition and a sentence in the unambiguous control condition (Table \ref{tab:gp_constructions}).
These sentences were read by a total 2,000 subjects using the SPR paradigm, which resulted in 47,695, 47,699, and 47,711 data points for the disambiguating critical word and its two spillover words respectively after data preprocessing.

\paragraph{Estimation of Surprisal-Based GPE.}
The LME models fit to SPR times of filler items were then used to generate predicted reading times (in ms) for sentences in the ambiguous condition and the unambiguous control condition.
Subsequently, the increase in the predicted reading times due to the increase in surprisal across conditions at the critical word and two spillover words was estimated as the magnitude of surprisal-based GPE.
To this end, another set of LME models that include a binary ambiguity condition, along with by-subject and by-item random intercepts was fit to the predicted reading times at each word for each construction.\footnote{Unlike \citet{huangetal24}, who estimated the GPE of all three constructions simultaneously through one regression model that used dummy coding for constructions, we fit separate LME models to each subset in order to not impose any dependence between each construction's estimated GPE. Additionally, the design of both the `filler item' and `increase in predicted reading times' LME models had to be simplified from the original specifications in \citet{huangetal24} due to convergence issues.}

\begin{figure*}[ht!]
    \centering
    \includegraphics[width=\textwidth]{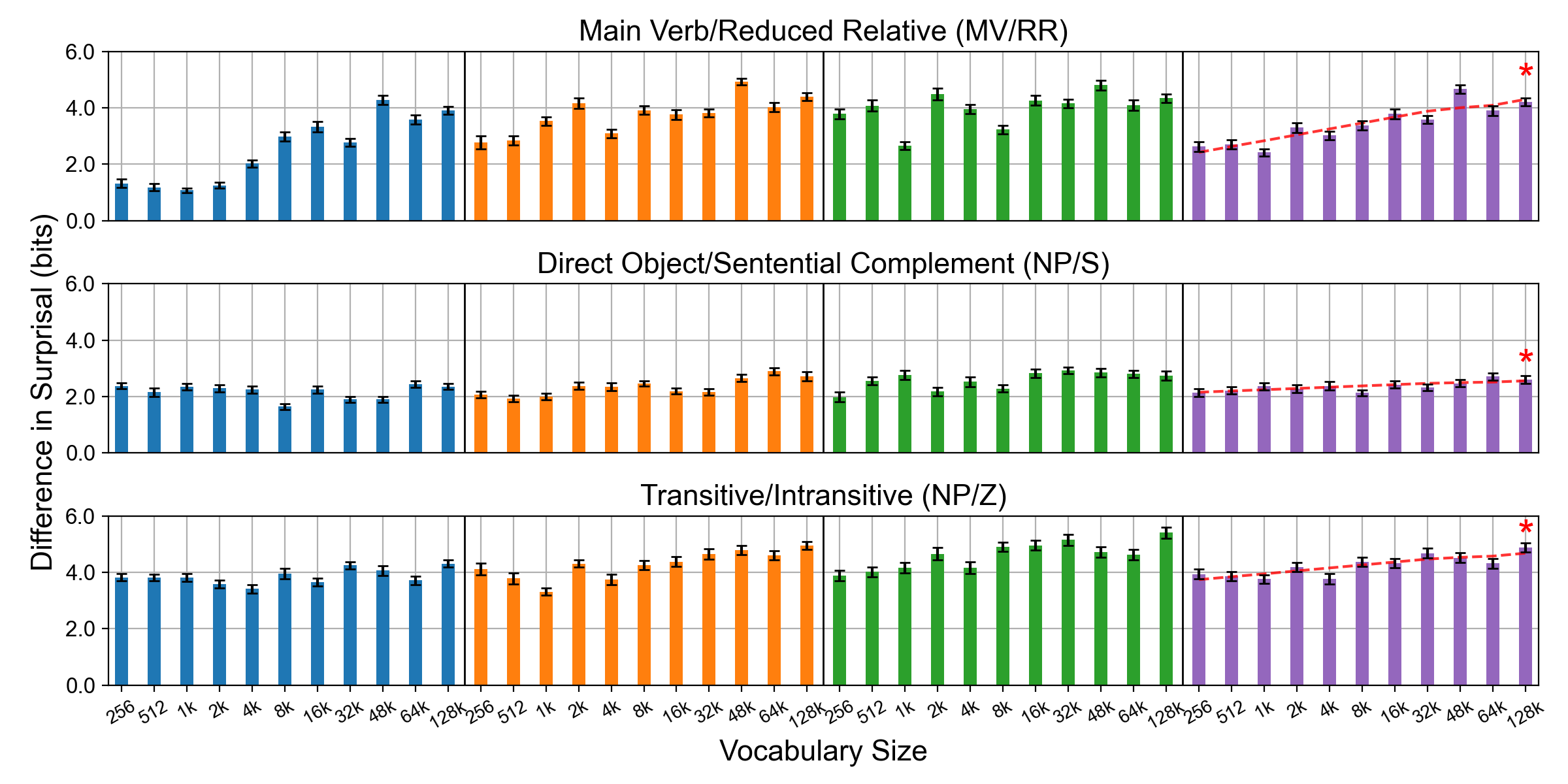}
    \includegraphics[width=0.67\textwidth]{figures/legend_size.png}
    \caption{Increase in LM surprisal of the critical word across conditions for the three garden-path constructions. Error bars denote 95\% confidence intervals. The red curve shows predictions from the line of best fit with $\log_2(|V|)$ as the independent variable. $*$: $p < 0.05$.}
\label{fig:gpe_surpdiff}
\end{figure*}

\subsection{Results}
The GPE estimated at the first spillover word\footnote{Effects are reported at the first spillover word as this region is where humans demonstrate the strongest GPE. See Appendix~\ref{sec:per_region} for the results at the critical word and the second spillover word.} in Figure \ref{fig:gpe_p1} again reveals an interaction between model size and token granularity, although the general trend is a lot less clear compared to Experiment~1.
The most notable trend is that the \textit{Small} LMs trained with larger vocabulary sizes demonstrate larger GPEs on the MV/RR condition compared to their counterparts trained with smaller vocabulary sizes.
This is consistent with the idea that having LMs treat words as independent symbols and learn from co-occurrences between them results in stronger representations of syntax.
In contrast, the \textit{Medium} LM trained with a vocabulary size of 1,000 appears to represent a peak in GPE across the three constructions, which demonstrates the opposite trend.

However, the raw difference in surprisal across conditions visualized in Figure \ref{fig:gpe_surpdiff} suggests that this peak is rather due to the difference in the estimated surprisal-reading time linking functions (i.e.~coefficients and spillover dynamics of various predictors); \textit{Large}, \textit{Medium}, and \textit{Small} LMs trained with larger vocabulary sizes generally show larger differences in surprisal at the critical word across all three conditions. 

Moreover, consistently with Experiment 1, the trend in estimated GPE or the difference in surprisal across conditions is the least clear in \textit{Large} LMs, although the models trained with larger vocabulary sizes tend to show larger differences in surprisal at critical words of NP/Z constructions.
Given sufficient model sizes, it may be the case that LMs learn predictions that are even more similar for short, isolated sentences like the garden-path stimuli compared to longer, more extended naturalistic corpora.
Finally, the manipulation of token granularity does not seem to alleviate the neural LMs' underestimation of human-like garden-path effects, which have been shown to be one or two orders of magnitude higher in previous work \citep{vanschijndellinzen21, arehallietal22, huangetal24}.

\section{Conclusion}
The influence of subword token granularity over which LMs are trained has been overlooked in cognitive modeling.
Nonetheless, this granularity directly encodes statistical information about word length and frequency into word probabilities.
Additionally, the granularity of tokens determines the collocational statistics between tokens within training corpora and ultimately impacts the quality of vector representations that are learned by LMs.

This work examines the influence of token granularity on the predictive power of LM surprisal on both naturalistic corpora and garden-path stimuli.
Experiments with naturalistic reading times reveal a substantial influence of token granularity both prior to and after LM training.
Tokens that are more fine-grained than contemporary standards resulted in LM surprisal that is most predictive, which suggests that the information about word length and frequency encoded by the tokenization process correlates with processing difficulty. 

In contrast, LMs trained on more coarse-grained tokens generally assigned higher surprisal to critical regions of garden-path constructions.
This may be due to the more direct word-to-word associations learned by LMs, which is facilitated by tokens that are more word-like.
As the critical word is identical across conditions in the garden-path stimuli, word length and frequency information appear to matter less in accounting for GPE.
Taken together, these results suggest a large role of token granularity on LM surprisal for cognitive modeling.

\section*{Acknowledgments}
We thank the ARR reviewers and the area chair for their helpful comments.
This work was supported by the National Science Foundation (NSF) grant \#1816891.
All views expressed are those of the authors and do not necessarily reflect the views of the NSF.
This work was supported in part through the NYU IT High Performance Computing resources, services, and staff expertise.

\section*{Limitations}
The influence of token granularity on the predictive power of surprisal identified in this work is supported by experiments using language models trained on English text and data from human subjects that are native speakers of English.
Therefore, it remains to be seen whether the findings will generalize to language models and data collected in other languages.
Additionally, although language models of multiple sizes were trained and evaluated in this work, models that are smaller or larger may yield different conclusions about the role of token granularity.
Finally, this work is concerned with the use of language models as cognitive models of human sentence processing, and therefore does not relate to their use in natural language processing applications.

\section*{Ethics Statement}
This work used data collected as part of previously published research \citep{futrelletal21, smithlevy13, copetal17, kennedyetal03, lukechristianson18, huangetal24}.
Readers are referred to the respective publications for more information on the data collection and validation procedures.
As this work focuses on studying the connection between conditional probabilities of language models and human sentence processing, its potential negative impacts on society appear to be minimal.

\bibliography{custom}

\appendix
\section{By-Corpus Regression Results} \label{sec:per_corpus}

$\Delta$LogLik for each LM surprisal evaluated in Experiment 1 on each reading time dataset is presented in Table \ref{tab:per_corpus}.

\aboverulesep=0ex
\belowrulesep=0ex
\newcolumntype{R}{>{$}r<{$}}
\begin{table*}[ht]
\scriptsize
\begin{tabular}{lRRRRRRRRRRR}
\toprule
    Model / $|V|$ & \text{NS} & \text{Brown} & \text{GECO}_\text{FP} & \text{GECO}_\text{GP} & \text{Dundee}_\text{SP} & \text{Dundee}_\text{FP} & \text{Dundee}_\text{GP} & \text{Provo}_\text{SP} & \text{Provo}_\text{FP} & \text{Provo}_\text{GP} & \text{Total} \\ \midrule
    \textit{Small} 256 & 1558.0 & 473.5 & 495.2 & 157.4 & \textbf{137.4} & \textbf{401.9} & 170.5 & 4.3 & 212.6 & 64.2 & 3675.0 \\
    \textit{Small} 512 & 1594.0 & 489.2 & 485.4 & 156.5 & 127.4 & 399.4 & 166.3 & 5.0 & 214.7 & 76.1 & 3714.0 \\
    \textit{Small} 1000 & 1666.0 & 498.5 & 534.6 & 171.8 & 127.7 & 399.8 & 178.1 & 4.3 & 222.7 & 72.0 & 3875.5 \\
    \textit{Small} 2000 & 1650.0 & 484.7 & 535.6 & 164.6 & 132.0 & 376.4 & 165.2 & 4.2 & 230.2 & 75.4 & 3818.3 \\
    \textit{Small} 4000 & \textbf{1681.0} & 484.7 & 551.5 & 170.1 & 128.4 & 378.6 & 168.0 & 5.3 & 211.3 & 75.3 & 3854.2 \\
    \textit{Small} 8000 & 1668.0 & \textbf{499.1} & \textbf{598.5} & 188.2 & 134.0 & 400.6 & 177.1 & 7.9 & 219.4 & 84.5 & \textbf{3977.3} \\
    \textit{Small} 16000 & 1599.0 & 480.3 & 552.8 & 169.4 & 136.3 & 399.4 & \textbf{181.6} & \textbf{10.0} & \textbf{251.5} & 89.3 & 3869.6 \\
    \textit{Small} 32000 & 1561.0 & 465.8 & 538.3 & 190.2 & 124.7 & 377.9 & 172.7 & \textbf{10.0} & 231.9 & \textbf{100.0} & 3772.5 \\
    \textit{Small} 48000 & 1510.0 & 462.1 & 550.5 & \textbf{193.3} & 126.4 & 359.6 & 160.0 & 8.3 & 210.6 & 92.8 & 3673.6 \\
    \textit{Small} 64000 & 1497.0 & 450.6 & 548.4 & 173.8 & 112.1 & 361.8 & 158.8 & 6.3 & 191.3 & 82.2 & 3582.3 \\
    \textit{Small} 128000 & 1556.0 & 440.6 & 523.6 & 181.0 & 116.4 & 345.9 & 151.8 & 9.7 & 185.1 & 86.1 & 3596.2 \\ \midrule
    \textit{Medium} 256 & 1582.0 & 470.8 & 538.5 & 180.4 & 124.6 & 361.8 & 152.9 & 9.8 & 192.3 & 80.0 & 3693.1 \\
    \textit{Medium} 512 & 1613.0 & 485.1 & 534.9 & 177.6 & 129.6 & 347.4 & 153.8 & 6.2 & 205.3 & 78.7 & 3731.6 \\
    \textit{Medium} 1000 & \textbf{1626.0} & \textbf{504.4} & 522.1 & 186.6 & 121.8 & 350.6 & 156.0 & 6.0 & 203.4 & 82.2 & \textbf{3759.1} \\
    \textit{Medium} 2000 & 1541.0 & 485.6 & 549.9 & 183.9 & 115.7 & 342.1 & 157.0 & 9.4 & 206.0 & 94.4 & 3685.0 \\
    \textit{Medium} 4000 & 1587.0 & 471.3 & 537.6 & 170.9 & 117.8 & 342.0 & 157.4 & 11.9 & 202.1 & 95.4 & 3693.4 \\
    \textit{Medium} 8000 & 1579.0 & 496.2 & \textbf{564.3} & 185.1 & 125.3 & 344.1 & 153.3 & 11.2 & \textbf{207.4} & 83.6 & 3749.5 \\
    \textit{Medium} 16000 & 1454.0 & 489.6 & 521.7 & 184.8 & 126.7 & 349.1 & 161.0 & 11.0 & 174.4 & 88.2 & 3560.5 \\
    \textit{Medium} 32000 & 1529.0 & 479.3 & 523.3 & 200.9 & \textbf{131.0} & \textbf{370.3} & \textbf{163.7} & 10.0 & 202.9 & \textbf{96.1} & 3706.5 \\
    \textit{Medium} 48000 & 1472.0 & 473.3 & 547.5 & 194.4 & 129.6 & 359.3 & 156.6 & 9.8 & 169.6 & 77.0 & 3589.1 \\
    \textit{Medium} 64000 & 1447.0 & 463.4 & 508.6 & 189.2 & 127.7 & 344.2 & 153.5 & 9.3 & 172.1 & 83.1 & 3498.1 \\
    \textit{Medium} 128000 & 1440.0 & 470.1 & 488.5 & \textbf{201.5} & 127.3 & 336.8 & 153.6 & \textbf{12.3} & 178.7 & 90.3 & 3499.1 \\ \midrule
    \textit{Large} 256 & 1553.0 & 473.0 & \textbf{534.4} & 184.0 & 120.6 & 326.6 & 147.3 & 9.4 & 175.0 & 75.7 & 3599.0 \\
    \textit{Large} 512 & 1531.0 & 482.5 & 503.1 & 179.5 & \textbf{136.3} & \textbf{336.9} & 153.9 & 7.6 & 177.3 & 83.1 & 3591.2 \\
    \textit{Large} 1000 & \textbf{1603.0} & 479.5 & 519.4 & 188.2 & 121.1 & 323.7 & 152.1 & 6.8 & 168.6 & 80.8 & \textbf{3643.2} \\
    \textit{Large} 2000 & 1558.0 & 479.6 & 513.4 & 165.5 & 122.7 & 326.3 & 147.5 & 8.5 & 172.9 & 77.3 & 3571.7 \\
    \textit{Large} 4000 & 1549.0 & 465.7 & 506.6 & 185.6 & 119.0 & 325.5 & 149.2 & 8.7 & 163.9 & 82.9 & 3556.1 \\
    \textit{Large} 8000 & 1536.0 & \textbf{490.5} & 511.5 & 185.4 & 119.8 & 335.3 & 153.9 & 10.6 & 178.7 & 85.2 & 3606.9 \\
    \textit{Large} 16000 & 1426.0 & 473.7 & 494.9 & \textbf{201.2} & 132.5 & 333.4 & \textbf{159.1} & 10.7 & \textbf{179.8} & \textbf{89.8} & 3501.1 \\
    \textit{Large} 32000 & 1428.0 & 487.8 & 498.5 & 190.0 & 124.4 & 332.5 & 155.7 & 9.2 & 176.9 & 84.0 & 3487.0 \\
    \textit{Large} 48000 & 1446.0 & 467.0 & 483.0 & 193.6 & 124.1 & 332.5 & 157.6 & \textbf{11.2} & 157.1 & 83.0 & 3455.1 \\
    \textit{Large} 64000 & 1410.0 & 466.3 & 491.2 & 188.2 & 124.1 & 332.9 & 150.7 & 8.1 & 146.5 & 82.6 & 3400.6 \\
    \textit{Large} 128000 & 1381.0 & 462.8 & 480.4 & 199.3 & 133.7 & 324.8 & 155.0 & 8.7 & 151.1 & 81.7 & 3378.5 \\ \midrule
    \textit{GPT-2 Small} & 1459.0 & 543.8 & 463.0 & 209.0 & 151.6 & 343.8 & 181.3 & 10.3 & 224.5 & 113.7 & 3700.0 \\ \bottomrule
\end{tabular}
\caption{$\Delta$LogLik of each LM surprisal evaluated in Experiment 1 on the exploratory partition of each reading time dataset. NS: Natural Stories.}
\label{tab:per_corpus}
\end{table*}

\section{GPEs Estimated at Other Word Regions} \label{sec:per_region}

\begin{figure*}[ht!]
    \centering
    \includegraphics[width=\textwidth]{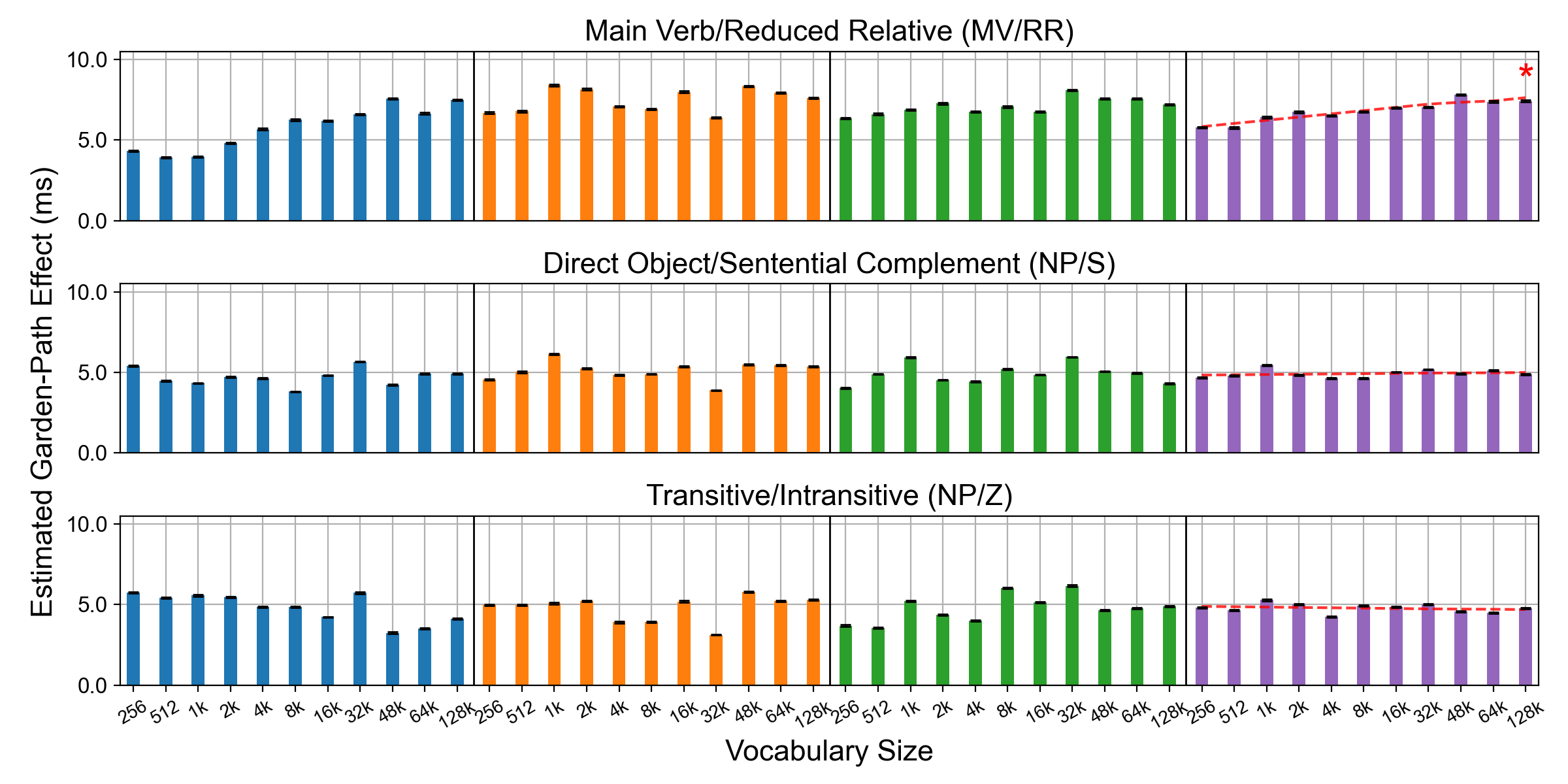}
    \includegraphics[width=0.67\textwidth]{figures/legend_size.png}
    \caption{Estimated garden-path effects at the critical word (\textit{deserved} in Table \ref{tab:gp_constructions}) for the three garden-path constructions using LM surprisal. Error bars denote 95\% confidence intervals. The red curve shows predictions from the line of best fit with $\log_2(|V|)$ as the independent variable. $*$: $p < 0.05$.}
\label{fig:gpe_p0}
\end{figure*}

\begin{figure*}[ht!]
    \centering
    \includegraphics[width=\textwidth]{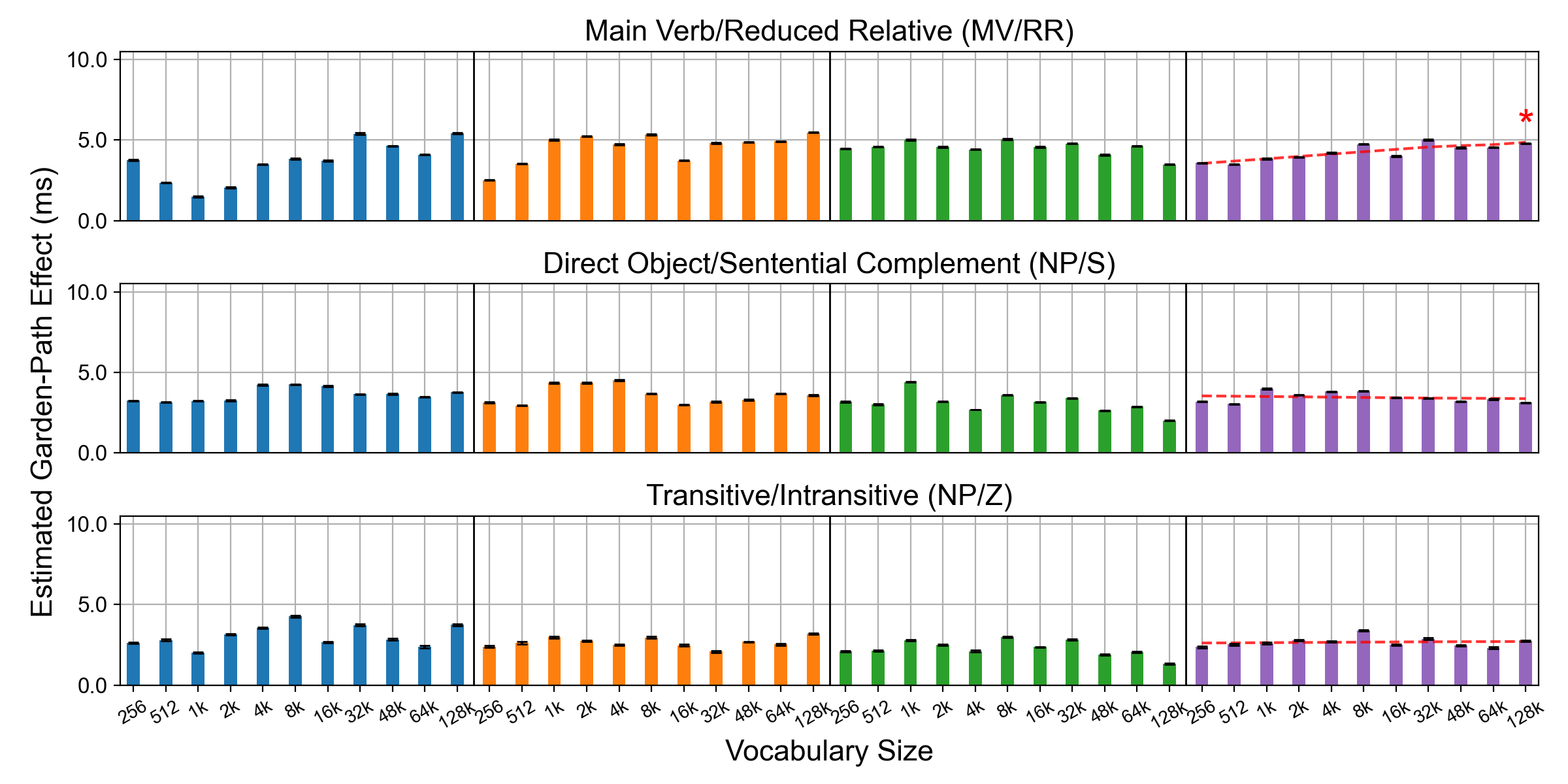}
    \includegraphics[width=0.67\textwidth]{figures/legend_size.png}
    \caption{Estimated garden-path effects at the second spillover word (\textit{investigation} in Table \ref{tab:gp_constructions}) for the three garden-path constructions using LM surprisal. Error bars denote 95\% confidence intervals. The red curve shows predictions from the line of best fit with $\log_2(|V|)$ as the independent variable. $*$: $p < 0.05$.}
\label{fig:gpe_p2}
\end{figure*}

The GPEs estimated at the critical word and the second spillover word are visualized in Figures \ref{fig:gpe_p0} and \ref{fig:gpe_p2} and respectively.

\end{document}